\documentclass[letterpaper, 10 pt, conference]{ieeeconf}  
\IEEEoverridecommandlockouts                               
\usepackage{graphicx}                       
\usepackage{graphics}                       
\usepackage{epsfig}                         
\usepackage[tight,footnotesize]{subfigure}  

\usepackage{amssymb,amsmath}
\usepackage{mdwmath}
\usepackage{mdwtab}
\usepackage{commath}                        
\usepackage{eqparbox}
\usepackage{amsmath,amsfonts,amssymb}

\newcommand{\ie}{{\em i.e.\ }}

\newcommand{\beq}{\begin{equation}}
\newcommand{\eeq}{\end{equation}}
\newcommand{\bear}{\begin{eqnarray}}
\newcommand{\bears}{\begin{eqnarray*}}
\newcommand{\eear}{\end{eqnarray}}
\newcommand{\eears}{\end{eqnarray*}}
\newcommand{\bdm}{\begin{displaymath}}
\newcommand{\edm}{\end{displaymath}}
\newcommand{\lba}{\left[\begin{array}}
\newcommand{\ear}{\end{array}\right]}

\usepackage{stfloats}                       

\usepackage{hyperref}

\usepackage{cite}                           

\usepackage[T1]{fontenc}

\usepackage{booktabs}						
\usepackage{multirow}						
\usepackage{tabularx}
\usepackage[table]{xcolor}

\newcommand{\squeezeup}{\vspace{-2.5mm}} 
\title{\LARGE \bf Online Robot Introspection via Wrench-based Action Grammars.}
\author{Juan Rojas, Shuangqi Luo, Dingqiao Zhu, Yunlong Du, Hongbin Lin,\\
Zhengjie Huang, Wenwei Kuang, and Kensuke Harada.\\
Guangdong University of Technology.\\
\thanks{J. Rojas and H. Lin is with the School of Electromechanical Engineering in the Guangdong University of Technology in Guangzhou, China.}%
\thanks{S. Luo, D. Zhu, Y. Du, Z. Huang, and W. Kuang are with the School of Software at Sun Yat Sen University in Guangzhou, China.}%
\thanks{Kensuke Harada is with the Intelligent Sys. Research Institute at AIST in Tsukuba, Ibaraki, Japan.}%
}
\begin{document}
\maketitle
\thispagestyle{empty}
\pagestyle{empty}
\bstctlcite{IEEEexample:BSTcontrol} 
\begin{abstract}
Robotic failure is all too common in unstructured robot tasks. Despite well-designed controllers, robots often fail due to unexpected events.
Robots under a sense-plan-act paradigm do not have an additional loop to check their actions. 
In this work, we present a principled methodology to bootstrap online robot introspection for contact tasks. In effect, we seek to enable the robot to recognize and expect its behavior, else detect anomalies. 
We postulated that noisy wrench data inherently contains patterns that can be effectively represented by a vocabulary. The vocabulary is obtained by segmenting and encoding data. And when wrench information represents a sequence of sub-tasks, the vocabulary represents a set of words or sentence and provides a unique identifier. The grammar, which can also include unexpected events, was classified both offline and online for simulated and real robot experiments. Multi-class Support Vector Machines (SVMs) were used offline, while online probabilistic SVMs were used to give temporal confidence to the introspection result. 
Our work's contribution is the presentation of a generalizable online semantic scheme that enables a robot to understand its high-level state whether nominal or anomalous. It is shown to work in offline and online scenarios for a particularly challenging contact task: snap assemblies. We perform the snap assembly in one-arm simulated and real one-arm experiments and a simulated two-arm experiment. The data set itself is also fully available online and provides a valuable resource by itself for this type of contact task. Our verification mechanism can be used by high-level planners or reasoning systems to enable intelligent failure recovery or determine the next most optimal manipulation skill to be used.
Supplemental information, code, data, and other supporting documentation can be found at \cite{2017IROS-Rojas-supplementalURL}.
\end{abstract}
\section{INTRODUCTION}\label{sec:Intro}
%
%
In autonomous scenarios, robotic failure is an undesirably frequent event. Despite well designed optimal controllers, robots can fail due to unexpected external events (internal too of course). Appropriately designed controllers give robots an action potential to reach set-points and reject disturbances; however, such controllers are unable to make sense of unexpected external events. We believe that due to a robot's lack of awareness about its own actions and the corresponding effect in the environment, they are unable to identify anomalous behavior and thus recover from it. From a different vantage point, a long-held paradigm in robotics has been the: ``sense-plan-act'' paradigm. We hold that such paradigm is limited in its ability to cope with uncertainty, so we wish to extend the latter to include a 4th element to form the: ``sense-plan-act-verify'' paradigm.
\begin{figure}
    \centering
        \includegraphics[scale=0.9]{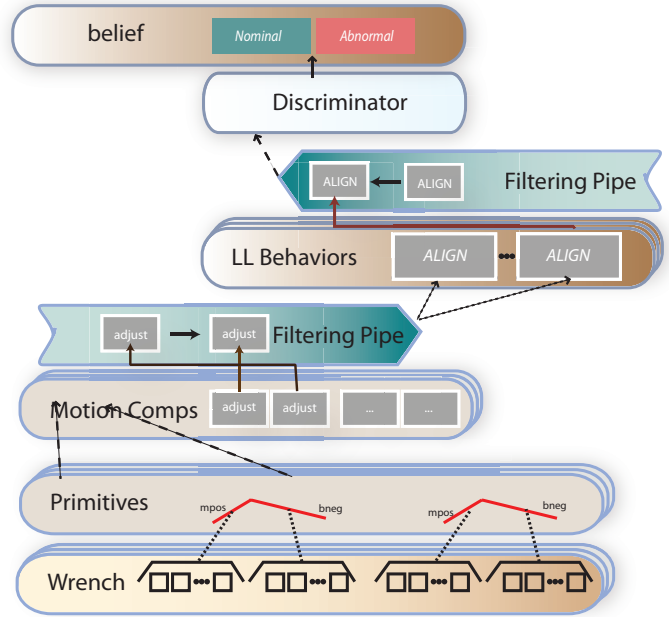}
        \caption{The online RCBHT process noisy wrench signals in fixed-length sets and pushes increasingly abstract encoded data up the system in a step-wise manner until the discriminator receives high-level action grammars that enable to infer the current robot state.}
        \label{fig:Taxonomy}
\end{figure}

In contact tasks, wrench signal interpretation is not as straight-forward as pose data. Wrench noise is not well approximated by Gaussian noise and may contain latent patterns that stem from the knowledge of an expert task programmer or human demonstrator. Such patterns may differ if the same task is executed by different agents in different ways. Given roughly similar signals, the goal of the verification step is to identify fundamental temporal patterns and model signal evolution to provide the necessary temporal introspection to the robot about its evolving high level state. If successful, a robot can use this information to reason about its next move: whether it is selecting the next skill to accomplish a task or recovering from abnormal behaviors (internal or external). Much work in the manipulation literature has gone into identifying robot skills that are flexible and reusable \cite{2015ICRA-Kroemer-TwrdsLearnHierSkillsMultiPhaseManip,2015IJRR-Niekum-LrnGrnddFiniteStateReprUnstrucDems}; less work has been done in the verification arena, where the robot is able to confirm nominal or abnormal behavior. Even more challenging is identifying not just abnormality but the type of abnormality that is experienced.
In verification, most work is divided into model-based techniques and data-driven techniques. If available, models of the system or the environment can be exploited to yield state estimates, though models are not always available due to system complexity \cite{2014IROS-Lowrey-ConsistentSensrFusionContactRichBehaviors,2016ICRA-Zhou-ConvexPolyForceMtnModel}. On the other hand, data-driven approaches collect data from one or more sensors and often use probabilistic \cite{2013IROS-DiLello-BayesianContFaultDetection,1998IJRR-Hovland-HMM_ProcessMonitorAsmbly} or machine learning concepts \cite{2010CASE-Rodriguez-FailureDetAsmbly_ForceSigAnalysis, 2011IROS-Rodriguez-AbortRetry, 2014ICRA-Rojas-EarlyFC} to estimate the task's state.
Data-driven techniques can be categorized as: (i) Discrete- \emph{vs.} Continuous-Event Evaluation and (ii) Low-Level State Estimation \emph{vs.} High-Level State Estimation. For discrete-event evaluations, contact points are evaluated as a contact sequence \cite{1998IJRR-Hovland-HMM_ProcessMonitorAsmbly, 2010CASE-Rodriguez-FailureDetAsmbly_ForceSigAnalysis, 2011IROS-Rodriguez-AbortRetry, 2015ICRA-Golz-TactileSensingLearnContactKnowledge}, whilst for continuous evaluations, it is the signal evolution that is modeled \cite{2013IROS-DiLello-BayesianContFaultDetection, 2013IJMA-Rojas-TwrdsSnapSensing}. For low-level state estimation, the output for event modeling is usually numeric \cite{2014IROS-Lowrey-ConsistentSensrFusionContactRichBehaviors,2016ICRA-Zhou-ConvexPolyForceMtnModel, 2015ICRA-Golz-TactileSensingLearnContactKnowledge, 2011IROS-Rodriguez-AbortRetry},\cite{2010CASE-Rodriguez-FailureDetAsmbly_ForceSigAnalysis, 2007Tro-Meeussen-CtctStSegPartFilt}, while for high-level state estimation it is usually semantic \cite{2013IJMA-Rojas-TwrdsSnapSensing}. The notion of a semantic representation for tasks has been used in medical robotics by using pose information from surgical robots to measure the skill with which a surgeon performs a surgery \cite{2013MICCAI-Ahmidi-StringMotifDescrToolMotion_SkillGestures}.

This work's contribution is a principled methodology to bootstrap online robot introspection for contact tasks through a continuous, data-driven, high level semantic state approach. In effect, we postulate that our approach can handle variants in signal uncertainty and effectively segment, encode, and classify (see Sec. \ref{sec:RCBHT} for details) the signal to yield an inference about the robot's current state. In fact, we use a particularly challenging assembly scenario: the multi-snap assembly of plastic parts (see Fig. \ref{fig:ExperimentalSetup} for details). Such assembly is characterized by high elastic forces during insertion and can be challenging to control. The current work differs from our previous efforts in a number of ways: (i) the ability to infer a state in the presence of either nominal or unexpected scenarios, (ii) the ability to do so online, (iii) robust testing by performing introspection in both real tasks (one-arm snap assemblies) and simulated tasks (one- and two-arm snap assemblies), and (iv) by making such data available to the public in an organized data-set .  Fig. \ref{fig:Taxonomy}, offers an overview of our online introspection method.

Our online segmenting and encoding methodology yield a vocabulary that represents fundamental temporal patterns in the wrench signal. These two processes are performed by the online Relative Change-Based Hierarchical Taxonomy (RCBHT) \cite{2013IJMA-Rojas-TwrdsSnapSensing}. The online RCBHT captures, during a time-window, relative change  in wrench data by fitting straight line regression segments to the signal, minimizing the effect of noise. The data is then encoded into a series of increasingly abstract layers through a small set of categorical labels yielding an action grammar for that segment of time and, until the end of the task. We assume nominal tasks are composed of a sequence of sub-tasks or phases \cite{2015ICRA-Kroemer-TwrdsLearnHierSkillsMultiPhaseManip}. For each sub-task the action grammar forms a set of words (sentence) that uniquely describes that phase. We perform the classification of sentences both offline and online. Multi-class Support Vector Machines (SVMs) were used offline, while online probabilistic SVMs were are used to give temporal confidence to the introspection result. We compare their discriminating efficacy in three robot  tasks: (i) one-arm real-robot assemblies, (ii) one-arm simulated assemblies, and (ii) two-arm simulated assemblies.

The contribution of our work is the presentation of a generalizable online semantic scheme that enables a robot to understand its high level state whether nominal or abnormal. It is shown to be robust by reporting efficacy in introspection in offline and online scenarios and in the use of 3 data-sets (real and simulated). Particularly, the introspection is done in a challenging contact task: snap assemblies. The data set itself is also fully available online and provides a valuable resource (it seems the only data set of this kind) for this type of task. This verification mechanism can be used by high-level planners or reasoning systems to enable intelligent failure recovery or determine the next most optimal manipulation skill to be used.

Our results show that offline introspection had very competitive mean value ranges from $89\%$-$100\%$; while online introspection had very high accuracies from $95\%$-$100\%$ and overall confidence levels from $81\%$-$84\%$. The more surface contact during a task the harder it was to have high confidence levels.

The advantage of this introspection system is it's ability to expand to other sensory modes \cite{2016Humanoids-Rojas-StateEst_PosBased_Grammars}, while its semantic nature allows the system to suitably provide feedback to high-level planners or reasoning systems \cite{2015IJCAI-Konidaris_Lozano-SymbolAcquisitionForProgHLPlanning, 2013ExpBots-Matuszek-LearningParseNatlLangToCommands}.

The rest of the paper is organized as follows: Sec. \ref{sec:RCBHT} presents the online segmentation and encoding steps to wrench signals. Sec. \ref{sec:Classification} introduces offline and online classification algorithms. Sec. \ref{sec:Experiments} introduces the contact task experiments and results. Sec. \ref{sec:Discussion} discusses originality, strengths, weaknesses, and future work. Sec. \ref{sec:Conclusion} summarizes key findings.
\section{The Online Relative Change-Based Hierarchical Taxonomy}\label{sec:RCBHT}
The RCBHT is used in robot tasks composed of sub-tasks or phases. Subtasks can be generated through programming by demonstration, a finite state machine (FSM), or other means. The state segmentation time is assumed known. In \cite{2013IJMA-Rojas-TwrdsSnapSensing}, the offline RCBHT enables semantic encoding of low-level wrench data. The taxonomy is built on the premise that low-level relative-change patterns are classified through a small set of categoric labels in an increasingly abstract manner. 
A multi-layer behavior aggregating scheme is composed of three bottom-to-top increasingly abstract layers and two top-to-bottom layers. Starting from the bottom layer and going up we have the Primitive layer, the Motion Composition (MC) layer, and the Low-Level Behavior layer (LLB). Then, top-down, we have the Introspection layer and the Classifier layer. The taxonomy is illustrated in Fig. \ref{fig:Taxonomy}.

In general, the framework separates each of the six Force-Torque (FT) axes; where, the Primitive layer partitions data for each of the six axis into linear segments that roughly approximate the original signal. In each segment, we extract features and provide a gradient classification label. The second layer (MC layer), examines gradient labels of primitive ordered-pairs according to a gradient pattern classification criteria. The ordered pairs are then categorized into a higher abstraction set. The third layer (LLB layer) applies the same logic to motion compositions to produce another higher abstraction layer. The advantage of the increasingly abstract multi-layer system is two-fold: (i) an increasingly intuitive semantic representation of behaviors, which is more suitable for higher-level planning and reasoning processes. (ii) The dimensional space size decreases by a factor of $2^x$, where $x$ is each new layer, thus decreasing computational cost and increasing speed; and (iii) Noise in the system is increasingly filtered with each new layer. Additionally, the classification from the top-down approach enables inference about which grammars best encode the sub-task(s) a robot performs.

The offline taxonomy required an entire traversal of the data before generating data at each of the different layers. The online system combines a one-item-look-ahead approach with more accurate contextual computations to generate consecutive abstractions at the lower-layers that are used as soon as new ordered-pairs are identified. The primitive layers reads a fixed-length segment of wrench signals to produce a linear segment and a corresponding gradient classification label. Labels are input into a filtering pipe (see Sec. \ref{subsec:Refinement}). When no more reductions are possible, the layer is published to the MC layer. The latter waits for an ordered-pair before it abstracts and feeds to that layer's filter. Again a stream of incoming labels are processed by the filter until no more reduction can take place. The respective label again is published to the above layer. A similar process occurs in the LLB layer, but this time the published label is fed to the classifier. The taxonomy was built in matlab-ros. The following sub-sections retain key RCHBT knowledge and point key differences between the online and the offline approach. For more details see \cite{2013IJMA-Rojas-TwrdsSnapSensing}.

\subsection{The Primitive's Layer}\label{subsec:Primitives}
\begin{figure}[ht]
    \centering
        \includegraphics[scale=1]{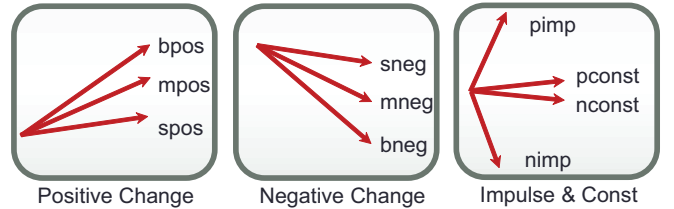}
        \caption{Gradient Classifications for Wrench Data in the Primitives Layer. There are 3 main groups: positive, negative, and const value gradients. To understand relative changes of magnitude, positive and negative gradients are divided into 4 regions: small, medium, large, and impulses.}
        \label{fig:Primitives}
\end{figure}

The Primitives layer partitions wrench data into linear data segments and classifies them according to gradient magnitude. Linear regression along with a correlation measure (the determination coefficient $R^2$) are used to segment data when a minimum correlation threshold is flagged.
Gradient classification for wrench data is fundamentally the set of three gradient value groups: positive, negative, and constant gradients. This simple classification is enough to capture relative change. Further, to understand relative magnitudes of change, positive and negative groups are subdivided into four ranges: small, medium, large, and impulses (very large). Contact between surfaces is characterized by abrupt changes in wrench signals almost approximating an (positive or negative) impulse, while near constant gradients are those where wrench change is trivial. To generalize gradient thresholds, we calibrate contextual information in a task to set the largest and near-constant gradient values to corresponding upper and lower boundary values (see \cite{2012ROBIO-Rojas-GradientOptimization} for more details). 
For the online version, linear regression is applied to empirically-set fixed-length segments of wrench data. The MC layer is triggered upon the generation of two linear segments and their classification as shown in  Fig. \ref{fig:Primitives}.
\subsection{The Motion Composition's Layer}\label{subsec:MCs}
\begin{figure}[ht]
    \centering
        \includegraphics[scale=1]{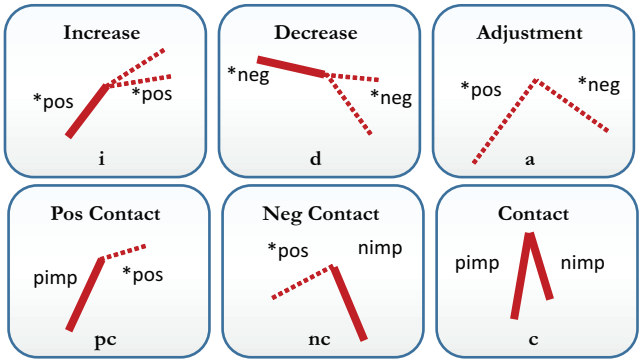}
        \caption{Illustration of possible generation of MCs. MCs are constructed by ordered-pairs of primitives. Six main groupings are shown. Each MC appears with its name, corresponding label, and ordered primitive pair. Dotted lines represent segments that can take primitives of different magnitudes, i.e. small-, medium-, or big-positive ones}.
        \label{fig:MotComps}
\end{figure}

The MC layer classifies ordered-pairs of primitives into seven categories: Adjustment, Increase, Decrease, Constant, Contact, and Unstable. These categories still represent positive, negative, and near-constant gradient groupings but also give rise to adjustments and unstable motions. Adjustments are primitive ordered-pairs that have a positive-negative or negative-positive transition. Adjustments represent quick ``back-and-forth" jerk action on the end-effector that are typical in alignment and insertion operations where force controllers minimize residual errors. Positive and negative gradients are grouped in this way to maximize the likelihood of grouping any such jerks regardless of slight variations in magnitude. For Increase, Decrease, and Constant categories they group contiguous (small-to-big) positive, (small-to-big) negative, and constant primitives respectively. For Contacts any positive or negative contact followed by any (small-to-big) negative primitive or (small-to-big) positive primitive yields a Contact, as well as a positive contact followed by a negative contact or vice-versa. These groupings along with their respective labels are illustrated in Fig. \ref{fig:MotComps}.
For the online version, an MC layer does not wait for the entire task's linear segments. Instead, as soon as a linear-segment pair is published, the layer works to produce an MC classification. That classification then goes into the filtering pipe. The filtering pipe will continue to take in MC labels until it can filter no more; at which time it publishes its result to the above layer.  
\subsection{The Low Level Behavior Layer}\label{subsec:LLBs}
The LLB layer classifies ordered-pairs of MCs into seven categories: Push, Pull, Fixed, Contact, Alignment, Shift, and Noise. The classification criteria is similar to the MC level but extends the definition of adjustments into increasingly stable adjustments (alignments) or increasingly unstable (shifts). Push and Pull group contiguous pairs of increase and decrease MCs respectively. Fixed and Contact group contiguous pairs of constant or contact MCs respectively. A major difference between the MC- and the LLB-level is the introduction of a shifting behavior `SH'. Shifts differ from alignments in that, whenever there are two contiguous adjustment compositions, if the second composite's amplitude is larger than the first, then it is a Shift. In effect, Alignments are adjustments that converge while Shifts become unstable over time. The visual representation of this layer is similar to Fig. \ref{fig:MotComps}, so it has been omitted for brevity.
\begin{figure*}
    \centering
        \includegraphics[width=1.0\linewidth]{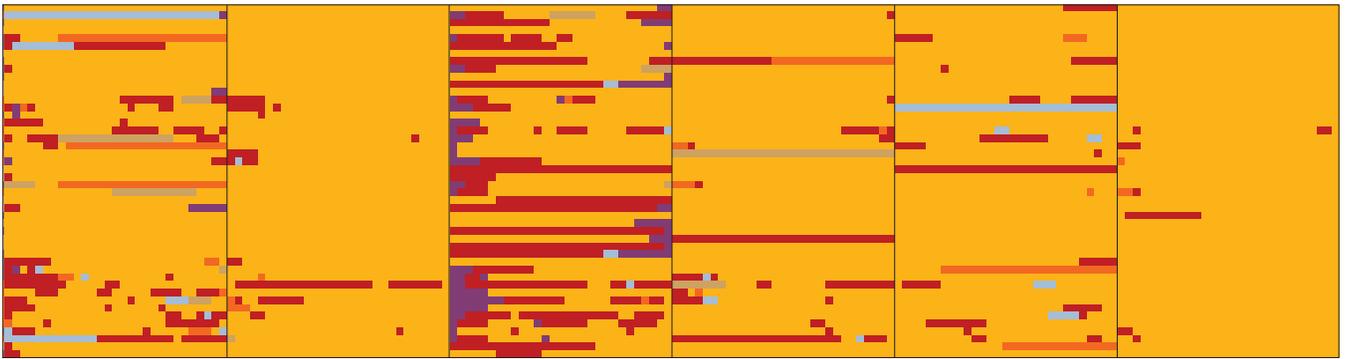}
        \caption{A color coded representation of the action grammar produced by LLBs for six FT axes stacked next to each other. For each stack, each column represents a different word. Each row represents a different trial. Patterns are evident at simple view across each FT axes. It is also visible that different axes contains very different types of information.}
        \label{fig:LLBcolorCodedMap}
\end{figure*}
For the online version, as before, the LLB layer does not wait for the entire tasks MC labels. Rather, as soon as an MC pair is published, the layer works to produce an LLB classification. The latter is fed to the filtering pipe, which again continues to take LLB labels until no more can be reduced. At this time, the filtered label is fed directly to the classifier. Fig. \ref{fig:LLBcolorCodedMap} shows a color coded representation of the action grammar produced by LLBs in offline mode for each of the six force-torque (FT) axes.
\subsection{Filtering}\label{subsec:Refinement}
The interleaving of filtering upon the production of classification labels along with subtle modifications to the filtering heuristics represent some of the major changes to the framework in the online formulation. At the onset, note that there is an embedded redundancy in the first three system layers. I.e. positive-gradient pairs become `Increase' MCs, which become `Push' LLBs. We have found that this grouping paired with filtering encode behaviors well. As mentioned earlier, filtering does not wait (as it did in the offline version) until all the labels in a layer are published; rather, it accepts input labels as soon as they are published by the previous layer.  The filter continues to accept more labels processes as many of them until no further reductions can be made. At this point, it publishes the output label to the next layer where the process repeats until it is fed to the classifier. This results in a sequential stair-like process of pipes that accept, accumulate, filter, and fire when they can process no more. 

Filtering is based on: (i) repeated behaviors and a (ii) time-amplitude context. The first category merges repeated behaviors. The second category merges negligible signals unto significant ones. The process was update to only merge such signals only if the amplitude value and the time duration of a neighboring label was 5 times (or more) larger and longer than the other one. Previously, we only compared amplitudes of the same size, and merged if the time duration was 5 times or more the neighboring one. In the offline version, a layer in the taxonomy would be filtered 2-3 times to significantly reduce the label number; in the online version we filter incoming labels as much as possible, and then start again (see Fig. \ref{fig:Taxonomy}.  For additional details on parameters and thresholds see our supplemental information listed in the Abstract.
\section{Classification Mechanisms}\label{sec:Classification}
As part of the top-bottom RCBHT scheme, the classification layer enables the robot to associate the an action grammar sentence with a particular sub-task. The classification mechanism has a number of underlying assumptions: (i) that the way the task is performed is fairly consistent, yet this does not rest importance to the approach. We think that for many tasks: from the way a human holds a tool to the way one dresses up, humans execute tasks in a similar fashion, according to way they learned. (ii) Classification is performed both offline and online for nominal and abnormal behaviors. (iii) State-transitions are provided. In our case, the controller provides them when empirical thresholds were met. Autonomous segmentation approaches can be found in: \cite{2015ICRA-Kroemer-TwrdsLearnHierSkillsMultiPhaseManip}, \cite{2015IJRR-Niekum-LrnGrnddFiniteStateReprUnstrucDems}. Based on these assumptions, the introspection approach seeks to identify the current executing robot-phase given some control goal. In this work we used a multi-class SVM classifier \cite{2004SnC-Smola-SVRTutorial} (see Sec. \ref{subsubsec:svm}) for offline classification and probabilistic SVMs for online classification. SVMs were selected as they are known for competitive classification of non-linear features while handling high dimensional data sets (ours will range in the order of $10^3$) well. For the online case, we used Wu's SVM probabilistic estimates for multi-class classification by pairwise coupling \cite{2004JMLR-Wu-ProbEst_MultiClass_ClassifPairwiseCoupl}, the standard option for such classification in Scikit-Learn \cite{2011JMLR-Pedregosa-scikit}.
\subsection{Support Vector Machines}\label{subsubsec:svm}
SVMs approximate a boundary to separate binary classes through a hyperplane for large feature spaces. The feature vector is used to learn a hyperplane: $\omega^T x − b=0$, where $\omega$ are the weights and $b$ is the bias from the zero point. In effect, the separation of training points from the hyperplane is the functional margin $\hat{y}(\mathbf{x})$ and can be modeled as:
\begin{equation}\label{eqtn:hyperplane}
    \hat{y}(\mathbf{x}) = sgn(f( \mathbf{x})) = sgn( \mathbf{\omega}^T \mathbf{x} + \hat{\omega}_0 )
\end{equation}
The signum function $sgn$ consists of the 2-tuple ${1,−1}$ for nominal and abnormal class labeling, and $\mathbf{x}$ is the input vector
for training and testing. The SVM optimizes the functional margin by maximizing the margin and ensuring that each point is on the correct side of the boundary, that is $f(x)y_i>0$ and the objective becomes:
\begin{equation}\label{eqtn:gamma}
  \max_{ \mathbf{\omega},\omega_0 } \min_{i=1,..,N} \frac{y_i( \mathbf{\omega}^T \mathbf{x} + \hat{\omega}_0 )}{ \|\mathbf(\omega)\| }.
\end{equation}
The larger the geometrical margin the more
accurate the classifier (for details of the implementation see Sec. \ref{subsec:classifier_setup}. For probabilistic SVM estimates, the pairwise coupling approach is used for multi-class classification. The later combines comparison for each pair of classes through linear systems \cite{2004JMLR-Wu-ProbEst_MultiClass_ClassifPairwiseCoupl}. For an observation $x$ and class label $y$, the pairwise class probabilities $r_{ij}$ of $\mu_{ij}$ are $P(y=i|y=i\mbox{ or } j,x)$. The $i^\text{th}$ and $j^\text{th}$ training set classes are used to compute a model to approximate $r_{ij}$ as an approximation of $\mu_{ij}$. Using all rij, $pi=P(y=i|x)$ is computed for $i=1,...,k$.
\section{Experiments}\label{sec:Experiments}
In this section we present the experimental setup and procedures. Three sets of experiments were conducted by the HIRO robot and the OpenHRP system during the offline stage: a one-arm simulated-robot snap assembly experiment, a one-arm real-robot snap assembly experiment, and a two-arm simulated-robot snap assembly. For the online stage, only the first two experiments were executed.
\subsection{Testbed Setup}\label{subsec:ExperimentalSetup}
\begin{figure}[t]
    \centering
        \includegraphics{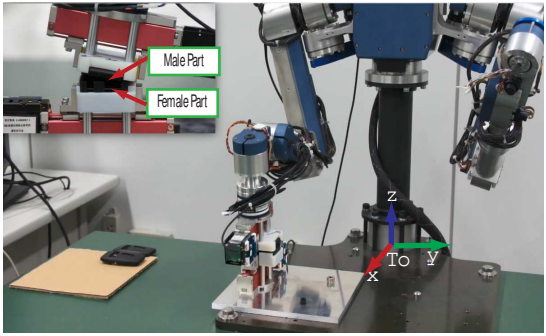}
        \caption{The HIRO humanoid robot is rigidly holding a male camera part. HIRO will run four sub-tasks to perform a snap assembly: guarded approach, a rotational alignment, a snap insertion, and a mating behavior.}
        \label{fig:ExperimentalSetup}
\end{figure}

HIRO, a 6 DoF dual-arm anthropomorphic robot is driven by stiff electric actuators. The robot uses a JR3 6DoF force-torque (FT) sensor rigidly attached on the wrist. A specially designed end-effector tool for rigidly holding a male and female plastic camera mold was also rigidly attached to the wrist. The robot is controlled through the OpenHRP environment \cite{2004IJRR:Kaneheiro:OpenHRP}. The camera parts are designed to snap into place. In fact, the male part consists of four snap beams (see Fig. \ref{fig:ExperimentalSetup}). A snap assembly strategy along with modular hybrid pose-force-torque controllers \cite{2013IJMA-Rojas-TwrdsSnapSensing} was used to pick up the part and then perform a set of four nominal sub-tasks: (i) a guarded approach to the female part, (ii) a rotational alignment procedure, (iii) a snap insertion where elastic forces can be very high, and (iv) a mating procedure that maintains the parts together before moving the arm away. Unexpected events that lead to failure usually occur during the initial contact point, \ie the localization of the parts is incorrect or a part has been moved for external reasons, failure could also happen during the insertion stage itself due to jamming or wedging.

The task was also simulated for a one-arm and a two-arm scenario. We used the OpenHRP's 3.0 simulation environment. The male and female camera parts were CAD rendered from the original ones. For the two-arm scenario, a lateral assembly was designed with the same strategy; however, the right arm functioned as an active arm while the left one functioned as a reactionary arm. The reactionary arm used force control to remain steady in spite of the right arm's push. In one-arm scenarios we segmented, encoded, and classified wrench data for only one-arm. But in the two-arm scenario we generated action grammars for both arms and performed the classification as a function of grammars in both arms. This too is the first time we report the encoding and classification of a dual-arm system. 

The tool center point (TCP) was placed on the point in the male camera where contact with the female part would occur first for a successful task. This point served as a global reference for the system and it was provided to the system \emph{a priori}.
The world reference frame was located at the manipulator's base. The TCP position and orientation were determined with reference to the world coordinate frame $To$. The force and torque reference frames were determined with respect to the wrist's reference frame.
\subsection{Classifier Setup}\label{subsec:classifier_setup}
Classifier setup is presented first for the offline scheme then for the online scheme. 
For offline experiments, we present the results in the following order: simulated one-arm task, real one-arm task, and simulated two-arm task. For online experiments, the results order is: simulated one-arm task and real one-arm task (we have yet to collect appropriate abnormal dual-arm task data, so it is not included online).
For both offline and online schemes, we organize our reporting in two ways: (i) abnormal classification: we infer whether probabilities of success or failure present in the data; and (ii) nominal state classification: we infer probabilities for each of the four sub-tasks.

Our goal is to classify grammars to aid the robot gain introspection about its current action. Nominal tasks consist of four sub-tasks: a guarded approach, a part's alignment, an insertion, and mating. Our training data is labeled for nominal and abnormal tasks and consists of all the listed sub-tasks unless an unexpected event occurs. We independently analyze each of the six wrench axes and each of the 3 RCBHT low-level labels. Fig. \ref{fig:LLBcolorCodedMap} illustrates the LLB labels for the insertion grammar for each of the six axes. Note that the grammar length varies for different axes in a given state. Thus, we perform a resampling step that computes the maximum number of labels for an RCBHT level and state across axes. Then, we extrapolate the labels for the rest of the axis to ensure equal grammar length. 5-fold cross-validation is used to train and validate all classifiers (offline and online).

Offline results are provided as the ${min,mean,max}$ classification accuracies at the end of training for the best kernel and the best parameter value(s). In our supplemental page \cite{2017IROS-Rojas-supplementalURL} however, we include graphs for all training showing accuracies over number of training examples, kernels, and parameters. Accuracy results are generated from validation trials. Additional results for all kernels and parameters are available in \cite{2017IROS-Rojas-supplementalURL}.

For the SVM classifier we tested linear, polynomial (poly), and radial basis function (rbf) kernels for multi-class support under a `one-versus-one' decision function shape. The penalty parameter $C$ was varied for powers of $1.10^{x}$, for $x=-5:1:4$ to examine possible overfitting. 

For the online scheme, we had the probabilistic version of the same SVM classifiers sample labels published by the lower RCHBT layers at rates of 2Hz, 10Hz, and 100Hz.  For online results, we measure multiple quantities: accuracy, per class probability, overall (non-zero) probabilities, and a confidence metric. The accuracy of the classifier may sometimes be 100, but we do not have an measure of confidence. The class probabilities can be considered a per class confidence measure. The overall probability is computed to help us understand the confidence of the (sub) task. The overall probability only averages probabilities of \textit{correct} classifications: $\frac{\sum_i^n P_i * b_i}{n}$, where $b_i=0$ for incorrect classifications, and $b_i=1$ for correct ones. With regards to confidence,  we compute metrics for a range of probabilistic thresholds $k=0.7:0.5:0.95$. Probabilities $(Pr<0.5)$ are inadmissible, $(0.5\leq Pr \leq k)$ are uncertain, and only those greater than the threshold $(Pr>k)$ are considered certain. Furthermore, we must also consider at what point in time we have confidence: early or late in the (sub) task? In general, we expect desirable confidence levels to occur towards the later part of the segment. If they happen earlier that is beneficial. We also need to look at the duration of a confidence level. What if we reach a confidence level only to lose it quickly thereafter? To this end, we use a metric $m$ that indicates the percentage of time (minimum, mean, and maximum results are reported) that confidence intervals were greater than our threshold in the latter one-third of a (sub) task. Note that a controller transition could be commanded to robot once the confidence threshold is reached. Thus even small numbers of the metric can be accepted. Nonetheless, the metric is reported to provide intuition about the duration of such confidence in the current (sub) task. See Table \ref{tab:svm_online_10Hz} for the results at the 0.70 confidence threshold. 
\subsubsection{Data-Set: One-arm Simulated Robot}
The simulated one-arm task consisted of 38 assembly trials with only nominal states and 38 trials with abnormal information, for a total of 76 trials for training for abnormality. For each nominal trial 4 samples (they correspond to the 4 sub-tasks derived from the FSM) were extracted yielding 152 samples. For each sample, features are generated by treating the six FT axes and the three lower-level RCBHT layers. This yields 2178 features for abnormality training and 1428 features when testing for nominal states. Mean accuracy results for abnormal and nominal state offline classification are presented in Table \ref{tab:svm_offline}. 

\subsubsection{One-arm Real Robot}
\begin{figure*}[th]
        \centering
            \subfigure{\includegraphics[scale=0.9]{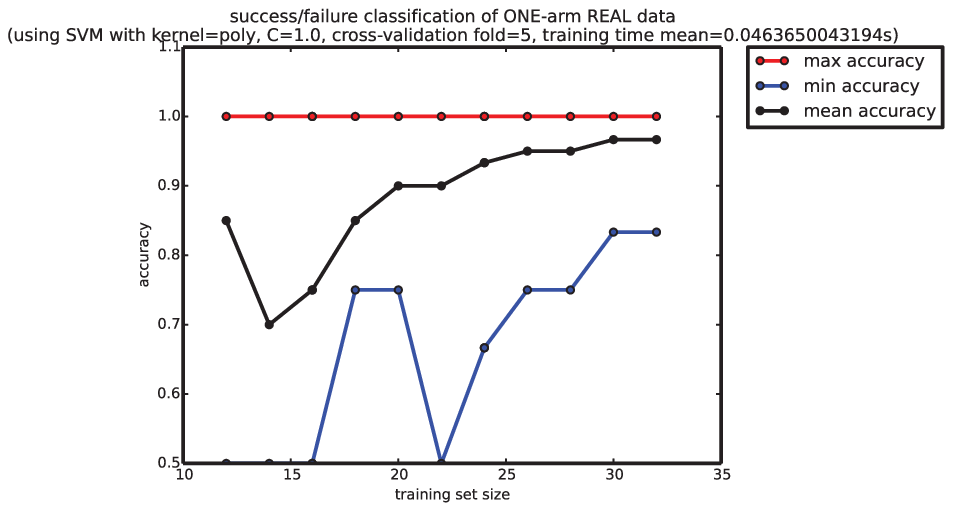}}
            \hspace{0.2cm}
            \subfigure{\includegraphics[scale=0.9]{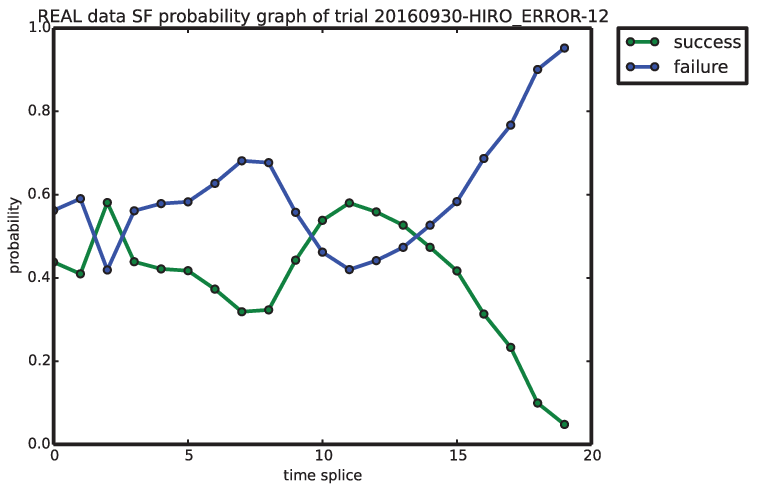}}\\
            \squeezeup
            \subfigure{\includegraphics[scale=0.9]{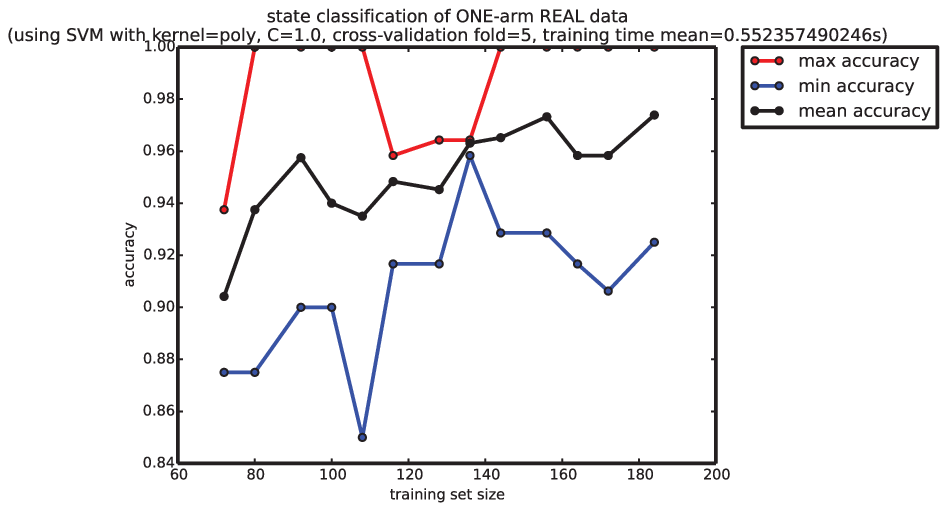}}
            \hspace{0.2cm}
            \subfigure{\includegraphics[scale=0.9]{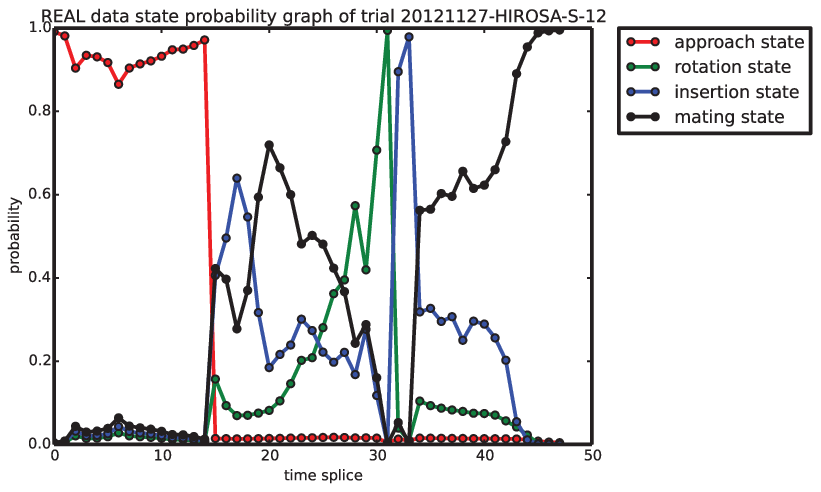}}\\
            \caption{Introspection results for offline (left column) and online (right columns) schemes. The first row represents inferences for task nominal vs abnormal results. The second row represents nominal states introspections.}\label{fig:results}
\end{figure*}

The real arm experiments consisted of 46 successful assembly trials and 16 failed assembly trails. We used 16 of 46 successful trials and all 16 failed trials to train a success/failure classifier totaling in 32 samples for abnormality training. For nominal state classification, we used all 46 successful trials yielding 184 samples. The feature length of the success/failure samples is 8352, while the feature length of state samples is 6204. We use 5-fold cross-validation to train these 2 classifiers. See row 1 in Fig. \ref{fig:results} for mean accuracy results for abnormal and nominal state offline inference respectively and row 2 for the online equivalent.
\subsubsection{Two-arm Simulated Robot}
The two-arm simulation consisted of 18 successful assembly trials but no failed trials. 72 state samples were used to train the nominal state classifier. The feature length for this experiment was 3786 (1782 from the left arm and 2004 from the right arm). Mean accuracy results for abnormal and nominal state offline classification can be seen in Table \ref{tab:svm_offline}.  
\subsection{Results} \label{subsec:Results}
Offline data shows that the SVM with a polynomial kernel and penalization parameter $C=1.0$ worked best across all data-sets and experiments. We use the mean accuracy to report results in this section. For the introspection of abnormality: the real HIRO one arm experiments achieved a mean of $94\%$ accuracy when using all training samples. The simulated HIRO arm data led to perfect classification from min to max values across a wide range of penalization parameters and for both the poly and linear kernels \cite{2017IROS-Rojas-supplementalURL}. For nominal state introspection, we got perfect classification for the dual-arm simulation, great classification at $97\%$ for the real robot, and $89\%$ for the one-arm simulated robot. 

Online data showed similarly competitive accuracies. $100\%$ for abnormal introspection and $95\%$ for nominal state inference. Overall confidence for abnormal inference stood at $81\%$ and $84\%$ for nominal state classification. Note that these results are unaffected by the selected threshold of 0.7. The threshold rather affects our metric which indicates the percentage of time we reach confidence in the last one-third of the (sub) task. At this confidence threshold level, we see that the guarded approach is the easiest to infer, the mating, but the alignment and insertion states which have the most instability are only confidently predicted towards the end of the nominal state. This is clearly seen in the 2nd-row, 2nd-column figure in Fig. \ref{fig:results}. An interesting fact that we noted was that in the first 21 trials of the real robot experiments we have one set of patterns, while in the rest of the trials (which were conducted at a later time) elicit a significant difference in patterns (\cite{2017IROS-Rojas-supplementalURL} contains the plots). The high accuracy classification remained despite the difference, however, the overall confidence and metric values used were affected by such difference.

\begin{table}[bt]
  \centering
  \caption{Offline SVM results for all data sets for best kernel (poly) and best penalty parameter C=1.0}
    \begin{tabular}{l|ccc}
    \textit{Abnormal Classification (Poly kernel)} & min & mean & max \\
    \midrule
    REAL\_ONE\_ARM & 0.83 & \textcolor[rgb]{ .773,  0,  .043}{0.97} & 1.00 \\
    SIM\_ONE\_ARM  & 1.00 & \textcolor[rgb]{ .773,  0,  .043}{1.00} & 1.00 \\
    \midrule
    \multicolumn{1}{l}{} \\
	\textit{Nominal State Classification} &&&\\
    \midrule
    REAL\_ONE\_ARM & 0.93 & \textcolor[rgb]{ .773,  0,  .043}{0.97} &  1.00 \\
    SIM\_ONE\_ARM  & 0.84 & \textcolor[rgb]{ .773,  0,  .043}{0.89} & 0.94 \\
    SIM\_TWO\_ARM  & 1.00 & \textcolor[rgb]{ .773,  0,  .043}{1.00} & 1.00 \\
    \end{tabular}%
  \label{tab:svm_offline}%
\end{table}%
\begin{table*}[bth]
  \centering
  \caption{Online SVM results for 0.70 confidence threshold sampled at 10Hz. Results include accuracy, overall probability, per class probability; min/mean/max count of confidence time-steps, avg experiment length, and min/mean/max metric values.}
    \begin{tabular}{lcl|c|cc|ccc|c|ccc}
    Type  & Thresh & Class & Acc   & OverallPr & ClassPr & minC  & meanC & maxC  & AvgLen & minM  & meanM & maxM \\
    \midrule
    REAL  & 0.7   & approach 	& \multirow{4}[2]{*}{95} & \multirow{4}[2]{*}{84} & 89    & 10.00 & 19.07 & 23.00 & 60.26 	& 0.50  & \textcolor[rgb]{ .753,  0,  0}{0.95} & 1.15 \\
    REAL  & 0.7   & rotation 	&       				 &       				  & 75    & 0.00  & 1.33  & 4.00  & 38.89 	& 0.00  & \textcolor[rgb]{ .753,  0,  0}{0.10} & 0.31 \\
    REAL  & 0.7   & insertion 	&       				 &       				  & 85    & 0.00  & 4.89  & 22.00 & 41.07 	& 0.00  & \textcolor[rgb]{ .753,  0,  0}{0.36} & 1.61 \\
    REAL  & 0.7   & mating 		&       				 &       				  & 86    & 0.00  & 39.20 & 108.00 & 173.96 & 0.00  & \textcolor[rgb]{ .753,  0,  0}{0.68} & 1.86 \\
    \midrule
    REAL  & 0.7   & success 	& 100    				& \multirow{2}[2]{*}{81} & 79    & 0.00  & 23.15 & 113.00 & 312.17 	& 0.00  & \textcolor[rgb]{ .753,  0,  0}{0.22} & 1.09 \\
    REAL  & 0.7   & Abnormal 	& 100   				&       				 & 83    & 0.00  & 7.50  & 19.00 & 127.33 	& 0.00  & \textcolor[rgb]{ .753,  0,  0}{0.18} & 0.45 \\
    \end{tabular}%
  \label{tab:svm_online_10Hz}%
\end{table*}%
\section{DISCUSSION} \label{sec:Discussion}
Our work shows that introspection can be bootstrapped under a segment, encode, and classify scheme in offline and online modes for challenging contact tasks in a variety of robotic tasks. High accuracy was achieved to detect nominal and abnormal situations and in online cases overall confidence levels above $80\%$ were achieved.
Our work presents very competitive results compared to those of similar works. DiLello \textit{et. al} \cite{2013IROS-DiLello-BayesianContFaultDetection}, classified nominal contacts tasks with $97.5\%$ accuracy and abnormal states with accuracies ranging from $80\%$ to $90\%$. Ahmidi \textit{et. al} \cite{2013MICCAI-Ahmidi-StringMotifDescrToolMotion_SkillGestures} classified surgical data through position-based grammar. Her results ranged from $75.2\%$ to $82.16\%$. 

The next step is to enact online decision systems that recover from failure gracefully. We also wish to implement similar work using Bayesian non-parametric models to remove the parametric dependence of our segmentation-encoding approach. Finally, we also wish to explore the use of online decision making systems in human-robot collaboration and so as to help a robot anticipate human behavior and provide better service.
\section{CONCLUSION} \label{sec:Conclusion}
A generalizable online robot introspection scheme for nominal or anomalous events in robot contact tasks was presented. A challenging contact task, the snap assembly, is attempted in one-arm simulated, one-arm real, and two-arm simulated experiments. Probabilistic multi-class SVM is used to provide confidence information about the robot's behavior. All data is also fully available online and provides a valuable resource by itself for this type of contact task. Such introspection mechanism can be used by high-level planners or reasoning systems to enable intelligent failure recovery or determine optimal subsequent manipulation skills to be used.
\section{Acknowledgments} \label{sec:Acknowledgments}
This work is supported by the grants: ``Major Project of the Guangdong Province Department for Science and Technology (2014B090919002), (2016B0911006).''
\bibliographystyle{IEEEtran}
\bibliography{IEEEabrv,Xbib}
\end{document}